\documentclass[10pt,twocolumn,letterpaper]{article}

\usepackage{cvpr}
\usepackage{times}
\usepackage{epsfig}
\usepackage{graphicx}
\graphicspath{{./img/}}
\usepackage{amsmath}
\usepackage{amssymb}
\usepackage[breaklinks=true,bookmarks=false]{hyperref}

\cvprfinalcopy

\begin{document}

\title{ComboGAN:\\Unrestrained Scalability for Image Domain Translation}

\author{Asha Anoosheh \\
Computer Vision Lab \\
ETH Z{\"u}rich \\
{\tt\small ashaa@ethz.ch}
\and
Eirikur Agustsson \\
Computer Vision Lab \\
ETH Z{\"u}rich \\
{\tt\small aeirikur@ethz.ch}
\and 
Radu Timofte \\
ETH Z{\"u}rich \\
Merantix GmbH \\
{\tt\small timofter@ethz.ch}
\and 
Luc Van Gool \\
ETH Z{\"u}rich \\
KU Leuven \\
{\tt\small vangool@ethz.ch}
}

\maketitle

\begin{abstract}
This year alone has seen unprecedented leaps in the area of learning-based image translation, namely CycleGAN, by Zhu~\etal. But experiments so far have been tailored to merely two domains at a time, and scaling them to more would require an quadratic number of models to be trained. And with two-domain models taking days to train on current hardware, the number of domains quickly becomes limited by the time and resources required to process them. In this paper, we propose a multi-component image translation model and training scheme which scales linearly - both in resource consumption and time required - with the number of domains. We demonstrate its capabilities on a dataset of paintings by 14 different artists and on images of the four different seasons in the Alps. Note that 14 data groups would need $(14$ choose $2) =$ 91 different CycleGAN models: a total of 182 generator/discriminator pairs; whereas our model requires only 14 generator/discriminator pairs.
\end{abstract}

\section{Introduction}

In such a short amount of time, we have come such a long way in the field of image domain adaptation and style transfer, with projects such as \cite{CYCLEGAN}, \cite{NVIDIA}, \cite{CYCADA}, \cite{PIX2PIX}, \cite{ADOBE}, \cite{STYLE}, and more paving the way. The first four are of particular interest as they do not simply transfer style in terms of texture and color, but in terms of semantics, and they maintain realism in their results. But they must be trained on examples from two specific domains, whereas the other two do not. The first three are even more noteworthy for being able to do this without any supervision between the two domains chosen - no pairing of matching data. Now what if we want to go beyond two domains?

Naively, we could train a model for each pair of domains we desire. With $n$ domains, this leads to $n \choose 2$ = $\Theta(n^2)$ models to train. In this work, we approach this problem by dividing each model into two parts: one that handles the conversion of one domain into a common representation and one that converts common representations into that domain. Having one of these pairs per domain allows us to mix-and-match by obtaining a common representation for any image and translating it to any other domain. All the while, the number of models increases linearly with the number of domains, as does the required training time.

\subsection{Generative Adversarial Models}

Creating a generative model of natural images is a challenging task. Such a model needs to be able to capture
the rich distributions from which natural images come from. Generative Adversarial Networks (GANs)~\cite{GAN}
have proven to be excellent for this task and can produce images of high visual fidelity. GANs, by default, consist of a pair of models (typically neural networks): a generator $G$ and a discriminator $D$. $D$ is trained to estimate the probability that a sample $x$ comes from a true training data distribution $p(x)$, while simultaneously $G$ is trained turn vector $z$ sampled from its own prior distribution $p(z)$ into $x'$, in order to maximize the value $D$ outputs when fed $x'$. $G$'s outputs receiving higher score by $D$ implies the distribution $G$ learns nears the true distribution $p(x)$.

This training procedure is referred to as adversarial training and corresponds to a Minimax game between $G$ and $D$. Training itself is executed in two alternating steps; first D is trained to distinguish between one or more pairs of real and generated samples, and then the generator is trained to fool D with generated samples. Should the discriminator be too powerful at identifying real photos to begin with, the generator will quickly learn low-level "tricks" to fool the discriminator that do not lie along any natural image manifold. For example, a noisy matrix of pixels can mathematically be a solution that makes $D$ produce a high real-image probability. To combat this, the training is often done with a small handful of examples per turn, allowing $G$ and $D$ to gradually improve alongside each other. The optimization problem at hand can be formulated as:

\begin{equation} \label{eq:gan_loss}
\min_{G} \max_{D} \mathbb{E}_{x} [log D(x)] + \mathbb{E}_{z} [log(1 - D(G(z))]
\end{equation}

\noindent GANs have found applications in many computer vision related domains, including image super-resolution \cite{SUPERRES} and style transfer \cite{STYLE}.

\subsection{Related Works}

Many tasks in computer vision and graphics can be thought of as translation problems where an input image
$a$ is to be translated from domain $A$ to $b$ in another domain $B$. Isola~\etal~\cite{PIX2PIX} introduced an image-to-image translation framework that uses GANs in a conditional setting where the generator transforms images
conditioned on the existing image $a$. Instead of sampling from a vector distribution $p(z)$ to generate images, it simply modifies the given input. Their method requires image data from two domains, but it requires they be aligned in corresponding pairs.

Introduced by Zhu~\etal, CycleGAN~\cite{CYCLEGAN} extends this framework to unsupervised image-to-image translation, meaning no alignment of image pairs are necessary. CycleGAN consists of two pairs of neural networks, $(G, D_A)$ and $(F, D_B)$, where the translators between domains $A$ and $B$ are $G : A \rightarrow B$ and $F : B \rightarrow A$. $D_A$ is trained to discriminate between real images $a$ and translated images $F(a)$, while $D_B$ is trained to discriminate between images $b$ and $G(a)$. The system is trained using both an adversarial loss, as expressed in \eqref{eq:gan_loss}, and a cycle consistency loss expressed in \eqref{eq:cycle_loss}. The Cycle consistency loss is a way to regularize the highly unconstrained problem of translating an image one-direction alone, by encouraging the mappings $G$ and $F$ to be inverses of each other such that $F(G(a)) \approx a$ and $G(F(b)) \approx b$. However, here the traditional negative log-likelihood loss in \eqref{eq:gan_loss} is replaced by a mean-squared loss \eqref{eq:lsgan_loss} that has been shown to be more stable during training and to produce higher quality results~\cite{LSGAN}. The full CycleGAN objective is expressed:

\begin{multline} \label{eq:lsgan_loss}
\mathcal{L}_{GAN}(G,D_B,A,B) = \\
	\mathbb{E}_b [(D_B(b) - 1)^2] + \mathbb{E}_a [D_B(G(a))^2]
\end{multline}
\begin{multline} \label{eq:cycle_loss}
\mathcal{L}_{cycle}(G,F) = \\
	\mathbb{E}_a [||F(G(a)) - a||_1] + \mathbb{E}_b [||G(F(b)) - b||_1]
\end{multline}
\begin{multline} \label{eq:cyclegan_loss}
\mathcal{L}(G,F,D_A,D_B) = \lambda\,\mathcal{L}_{cycle}(G,F) \\
	+ \mathcal{L}_{GAN}(G,D_B,A,B) + \mathcal{L}_{GAN}(F,D_A,B,A)
\end{multline}

\noindent The reconstruction part of the cycle loss forces the networks to preserve domain-agnostic detail and geometry in translated images if they are to be reverted as closely as possible to the original image. Zhu \etal were able to produce very convincing image translations such as ones trained to translate between horses and zebras, between paintings and photographs, and between artistic styles.

Liu~\etal~\cite{NVIDIA} implemented a similar approach with UNIT, adding further losses to the intermediate activation results within the generator instead of purely on the final generated outputs. Using the CycleGAN architecture, they designate the activations from the central layer of the generators as the shared latent vectors. Using a variational-autoencoder loss, these vectors from both domains are pushed into a gaussian distribution. This is done in addition to the discriminator loss and cycle loss from CycleGAN, seemingly improving the image translation task over CycleGAN in cases with significantly varying geometry in the domains.

Lastly there is the concurrent work of StarGAN~\cite{STARGAN}, which aims to solve a similar problem as ours: scalability of unsupervised image-translation methods. StarGAN melds the generators and discriminators from CycleGAN into one generator and discriminator used in common by all domains. As such, the model can take as input any number of domains, though this requires passing in a vector along with each input to the generator specifying the output domain desired. Meanwhile the discriminator is trained to output the detected domain of an image along with a real/fake label, as opposed to simply the latter when each domain has its own discriminator. The results suggest having a shared model for domains similar enough to each other may be beneficial to the learning process. Nevertheless, this method was only applied to the task of face attribute modification, where all the domains were slight shifts in qualities of the same category of images: human faces.

\section{The ComboGAN Model}

\begin{figure}[ht]
\begin{center}
\fbox{\includegraphics[width=0.9\linewidth]{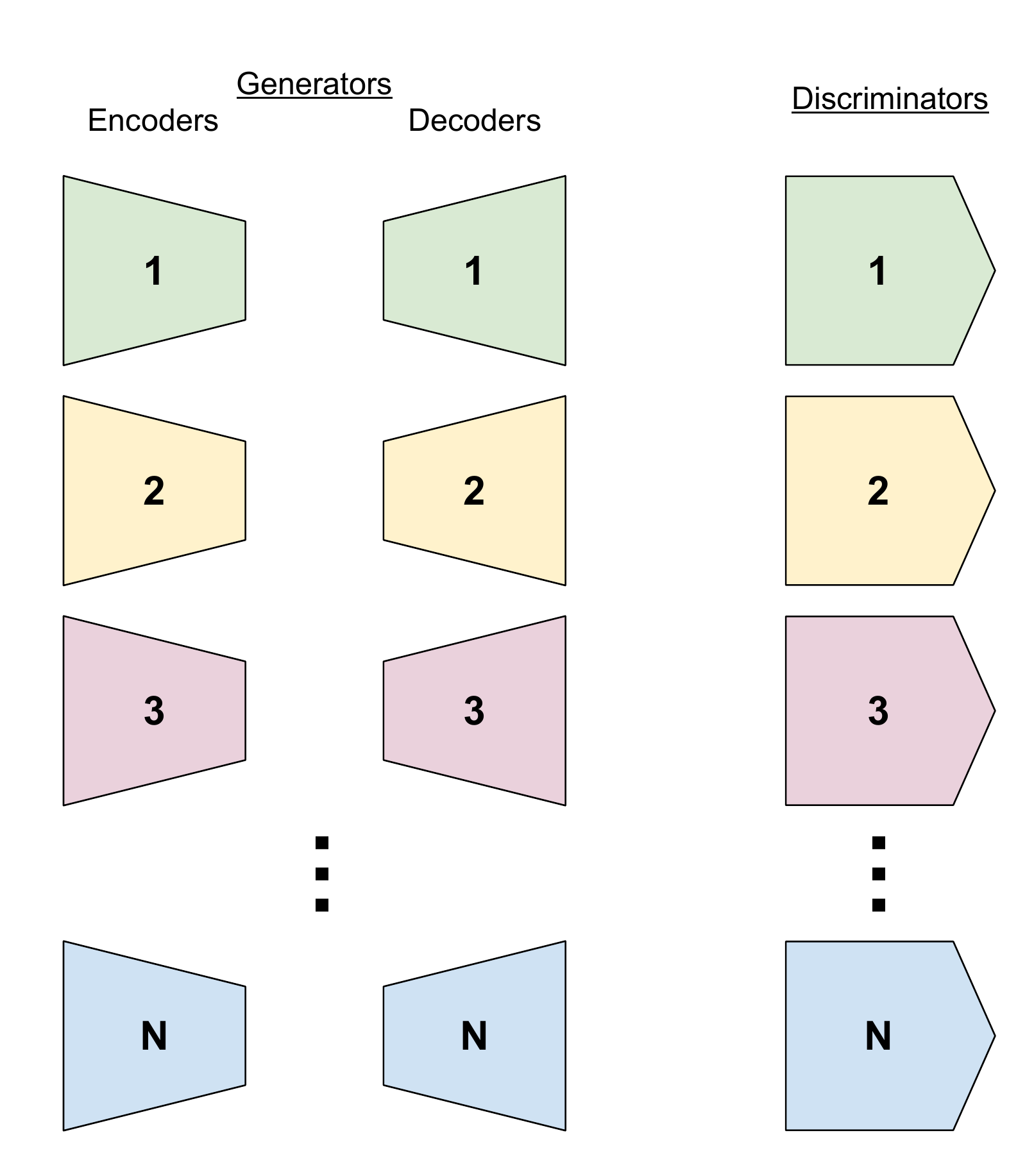}}
\end{center}
   \caption{Model design setup for N domains.}
   \label{fig:setup}
\end{figure}

\begin{figure*}[ht]
\begin{center}
\fbox{\includegraphics[width=0.8\linewidth]{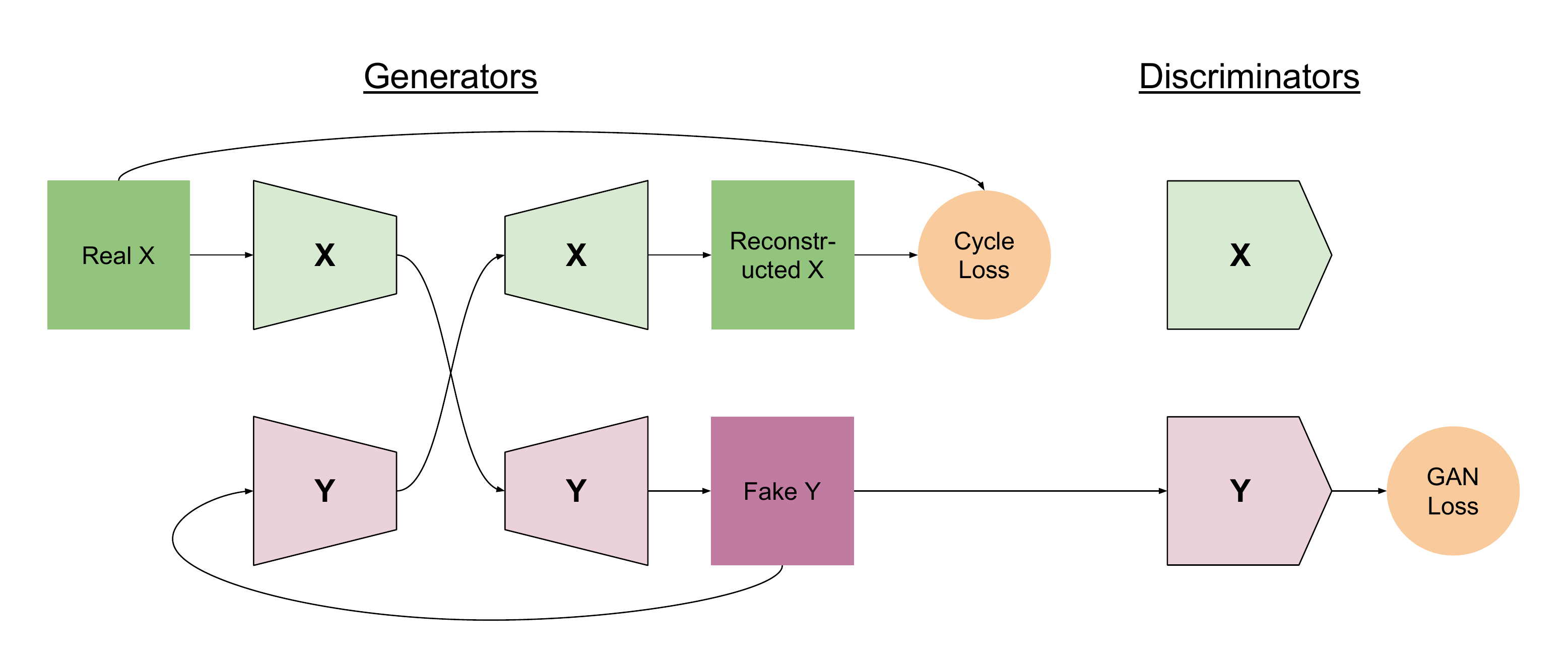}}
\end{center}
   \caption{Generator training pass for direction $X \rightarrow Y$, where $X, Y \in \{1,..,n\} : X \neq Y$ are randomly chosen from our $n$ domains at the start of every iteration. This pass is always repeated symmetrically for direction $Y \rightarrow X$ as well.}
   \label{fig:training}
\end{figure*}

\begin{figure}[h]
\begin{center}
\fbox{\includegraphics[width=0.94\linewidth]{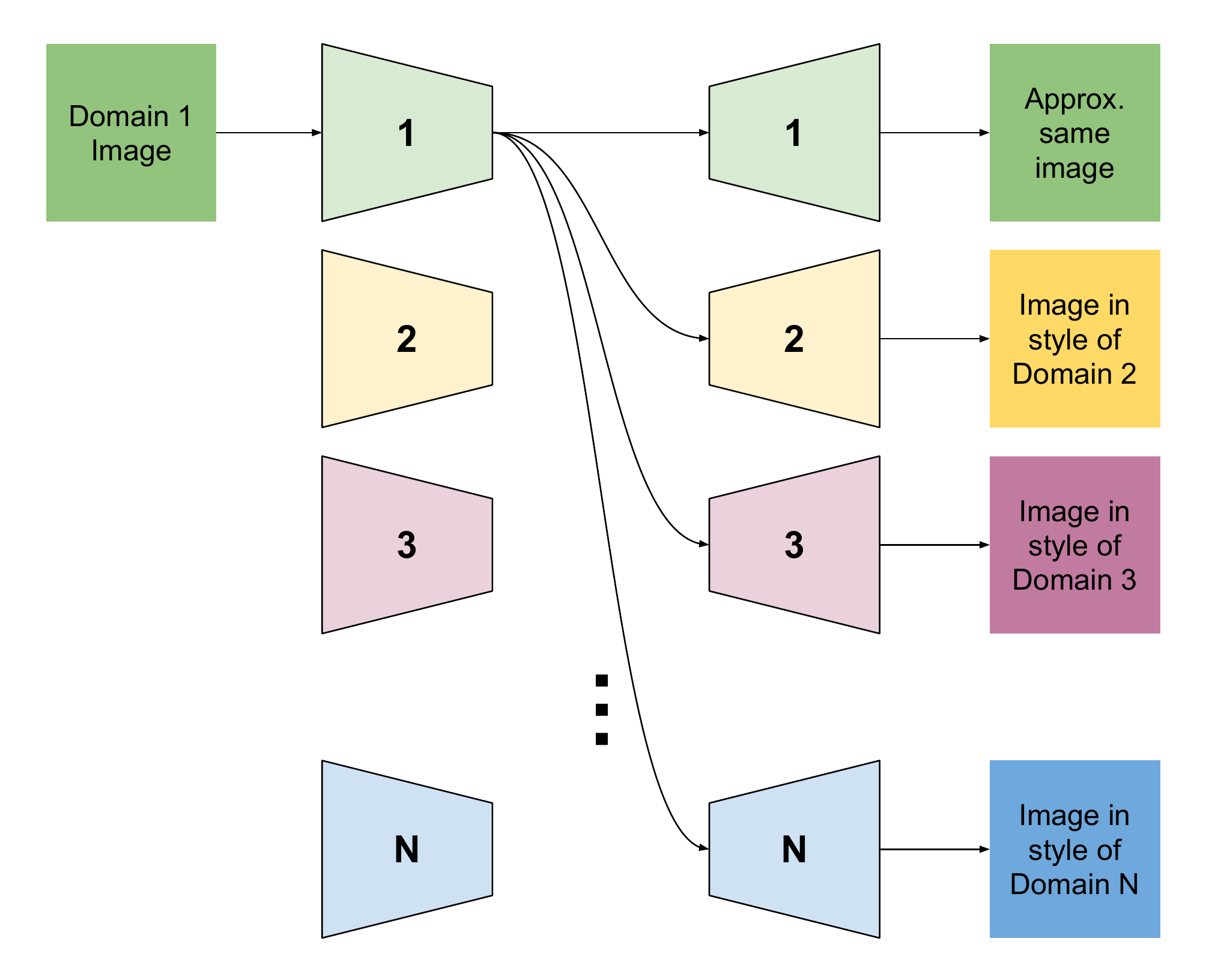}}
\end{center}
   \caption{Example inference functionality of translation from one domain to all others.}
   \label{fig:inference}
\end{figure}

\subsection{Decoupling the Generators}
The scalability of setups such as CycleGAN's is hindered by the fact that both networks used are tied jointly to two domains, one from some domain $A$ to $B$ and the other from $B$ to $A$. To add another domain $C$, we would then need to add four new networks, $A$ to $C$, $C$ to $A$, $B$ to $C$, and $C$ to $B$. To solve this issue of exploding model counts, we introduce a new model, ComboGAN, which decouples the domains and networks from each other. ComboGAN's generator networks are identical to the networks used in CycleGAN (see Appendix A for network specifications), yet we divide each one in half, labeling the frontal halves as encoders and the latter halves as decoders. We can now assign an encoder and decoder to each domain.

As the name ComboGAN suggests, we can combine the encoders and decoders of our trained model like building blocks, taking as input any domain and outputting any other. For example during inference, to transform an image $x$ from an arbitrary domain $X$ to $y$ from domain $Y$, we simply perform $y = G_{YX}(x) = Decoder_Y(Encoder_X(x))$. The result of $Encoder_X(x)$ can even be cached when translating to other domains as not to repeat computation.

With only one generator (an encoder-decoder pair) per domain, the number of generators scales exactly linearly with the number of domains, instead of $2{n \choose 2} = n(n-1) = \Theta(n^2)$. The discriminators remain untouched in our experiment; the number of discriminators already scales linearly when each domain receives its own. Figure \ref{fig:setup} displays our full setup.

\subsection{Training}
Fully utilizing the same losses as CycleGAN involves focusing on two domains, as the generator's cyclic training and discriminator's true/false-pair training are not directly adaptable for more domains. ComboGAN's training procedure involves focusing on 2 of our $n$ domains at a time. At the beginning of each iteration, we select two domains $X,Y \in \{1..n\}$ from our $n$ domains, uniformly at random. Then maintaining the same notation as CycleGAN in \eqref{eq:cyclegan_loss}, we set $A:=X$ and $B:=Y$ and proceed as CycleGAN would for the remainder of the iteration. Figure \ref{fig:training} shows one of the two forward passes in a training iteration. The other half is simply the symmetric mirroring of the procedure for the other domain, as if the two were swapped.

Randomly choosing between two domains per iteration means we should eventually cover training between all pairs of domains uniformly. Though of course the training time (number of iterations) required must increase as well. If training between two domains with CycleGAN requires $k_2$ iterations, then with $n$ domains, $n \choose 2$ CycleGAN setups would require $k_2 {n \choose 2}$ iterations to complete. In our situation, we instead keep the training linear in the number of domains, since the number of parameters (weights) in our model increases linearly with the number of domains, as well. We desire each domain to be chosen for a training iteration the same number of times as in CycleGAN. Note that it will not be the same number of times that a given pair is chosen, as achieving that would just require the same number of iterations as the naive method; rather we only care about whether a domain is chosen to be trained alongside any other domain or not. We observe that since a domain $X$ is chosen in each iteration with probability $\frac{2}{n}$, during training it is chosen in expectation $ \frac{n}{2} \cdot k_n $ times. Requiring equality to the two-domain case $k_2$, we obtain $k_n = \frac{k_2}{2} n$, or $\frac{k_2}{2}$ iterations per domain, which proves satisfactory in practice.

As for the discriminators, training is the same as CycleGAN's. After each training iteration for two given generators, the two corresponding discriminators receive a training iteration, as well. For a single discriminator, a real image from its domain and a fake image intended for the same domain (the output of that domain's decoder) are fed to train the network to better distinguish real and fake images. This is done independently for both discriminators.

\begin{figure}[t]
\begin{center}
\includegraphics[width=0.9\linewidth]{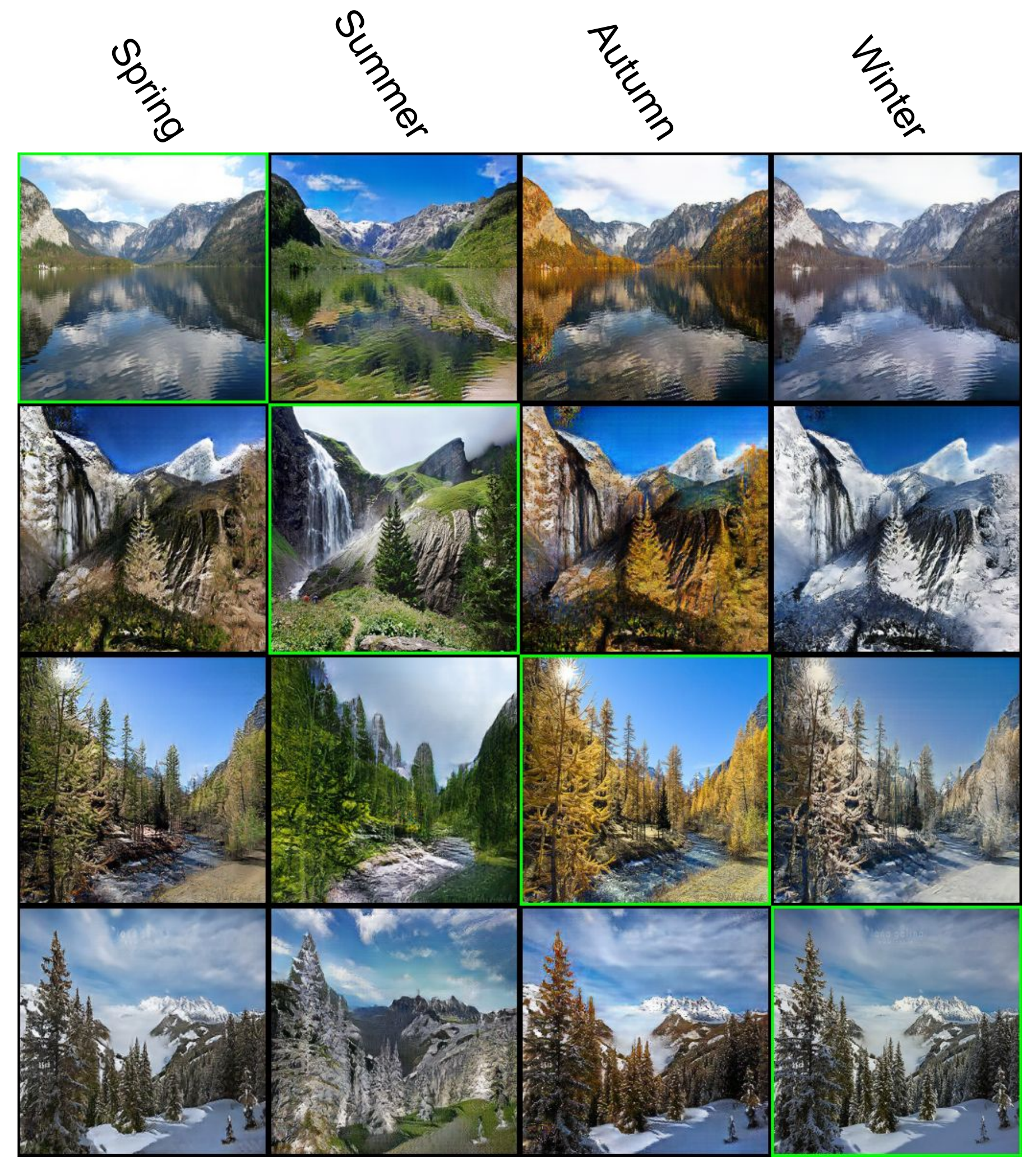}
\end{center}
   \caption{Validation results for pictures of the Alps in all four seasons. Original images lie on the diagonal.}
   \label{fig:alps}
\end{figure}

\begin{figure}[t]
\begin{center}
\includegraphics[width=0.9\linewidth]{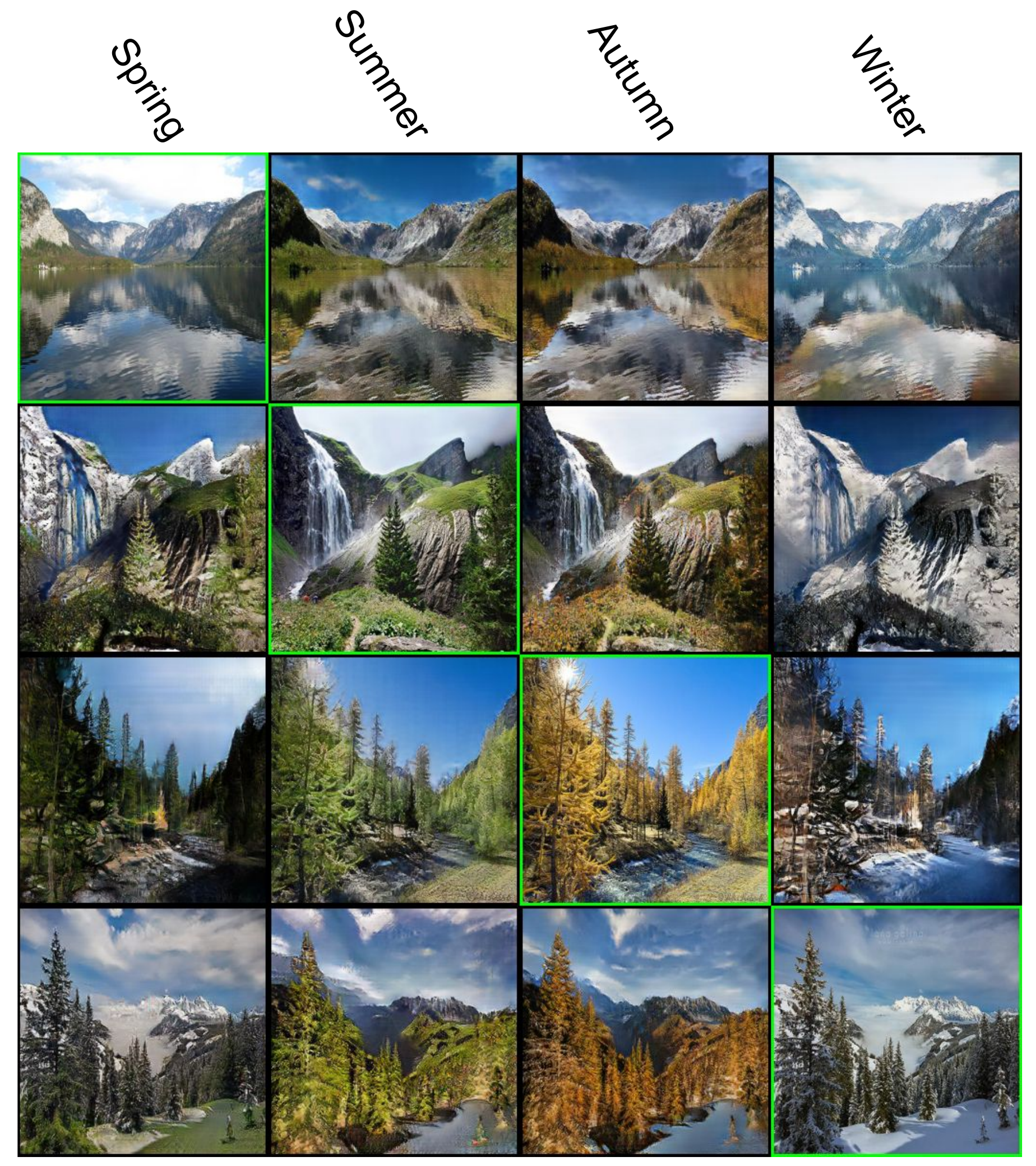}
\end{center}
   \caption{Same Alps images but from standard CycleGAN results instead. Original images lie on the diagonal.}
   \label{fig:vanilla_alps}
\end{figure}

\subsection{Relation with CycleGAN}
It is easy to see our changes only distinguish our model when more than two domains are present. For the case of two domains, our entire procedure becomes exactly equivalent to CycleGAN. Because of this, we consider ComboGAN a proper extension of the CycleGAN model that needs not change the underlying foundation.

In the case of more than two domains, for the end result to work as intended, it is implied the encoders must be placing input images into a shared representation, in which all inputs are equally fit for any domain transformation. Achieving this latent space suggests that the encoders learn to conceal qualities that make an image unique or distinguishable among the domains, with decoders re-filling them with the necessary detail that defines that domain's characteristics. As detailed in \cite{STENO}, cycle-consistent image translation schemes are known to hide reconstruction details in often-imperceptible noise. This could theoretically be avoided by strictly enforcing the latent space assumption with added losses acting upon intermediate values (encoder outputs) instead of the decoder outputs. ComboGAN's decoupled-generator structure allows for enhancements such as this, but for sake of direct comparison with CycleGAN, we omit tweaks to the objective formulation in this current experiment.

It should be noted though, that in the case of only two domains (and in CycleGAN), the concept of the images being taken to a shared latent space need not hold at all. In this situation, the output of an encoder is always given to the same decoder, so it will learn to optimize for that specific domain's decoder. In the case of more than two domains, the encoder output has to be suitable for all other decoders, meaning encoders cannot specialize.

\section{Experiments and Results}

\begin{figure*}[ht]
\begin{center}
\includegraphics[width=\linewidth]{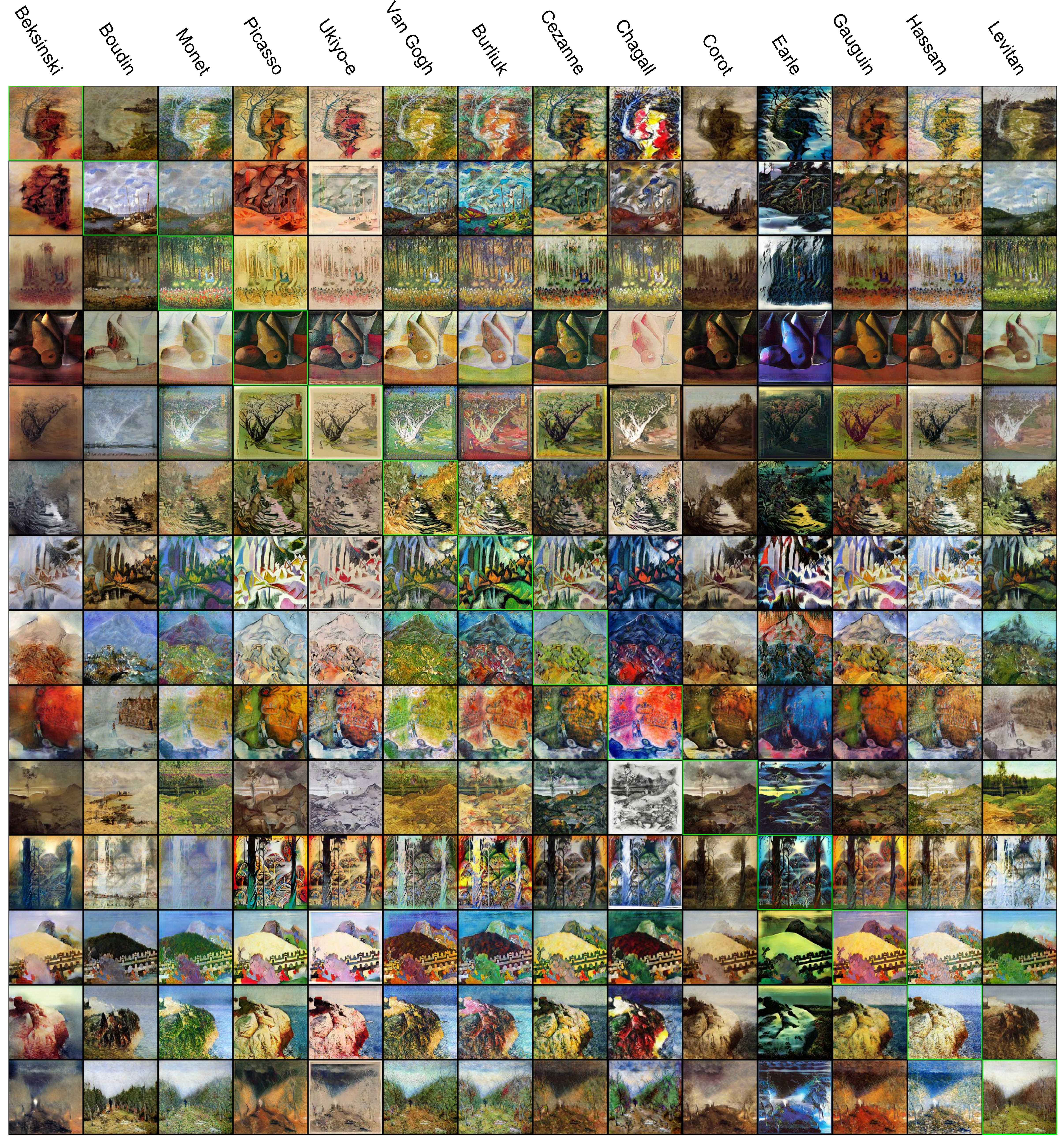}
\end{center}
   \caption{Validation results for our 14 painters. Original images lie on the diagonal.}
   \label{fig:paintings}
\end{figure*}

\begin{figure*}[ht]
\begin{center}
\includegraphics[width=\linewidth]{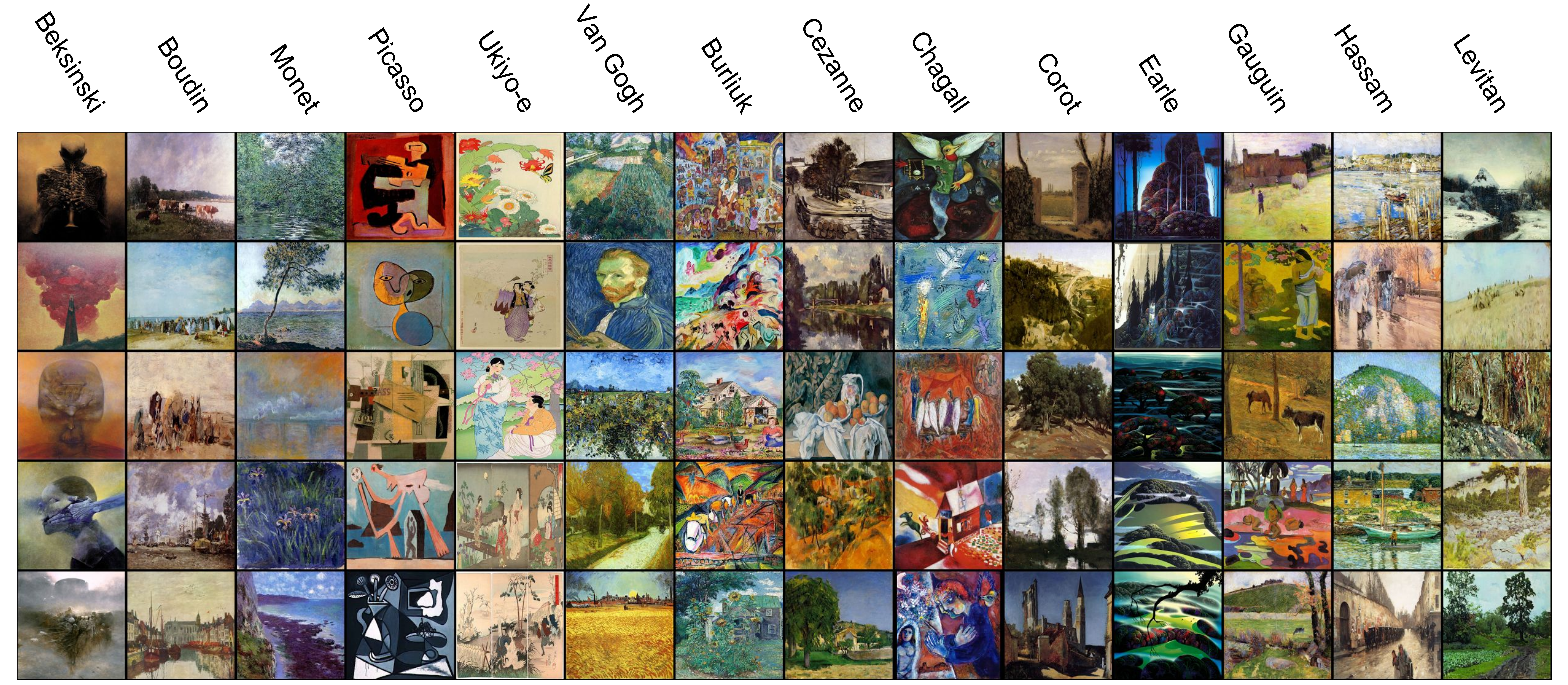}
\end{center}
   \caption{Sample artworks from our 14 painters to visually describe their styles.}
   \label{fig:samples}
\end{figure*}

\subsection{Datasets}
The first of two datasets used in this experiment consists of approximately 6,000 images of the Alps mountain range scraped from Flickr. The photos are individually categorized into four seasons based on the provided timestamp of when it was taken. This way we can translate among Spring, Summer, Autumn, and Winter.

The other dataset is a collection of approximately 10,000 paintings total from 14 different artists from Wikiart.org. The artists used are listed alphabetically: Zdzislaw Beksinski, Eugene Boudin, David Burliuk, Paul Cezanne, Marc Chagall, Jean-Baptiste-Camille Corot, Eyvind Earle, Paul Gauguin, Childe Hassam, Isaac Levitan, Claude Monet, Pablo Picasso, Ukiyo-e (style, not person), and Vincent Van Gogh.

\subsection{Setup}
All images in our trials were scaled to 256x256 size. Batches are not used (only one image per input), random image flipping is enabled, random crops are disabled, and dropout is not used. Learning rate begins at 0.0002 for generators and 0.0001 for discriminators, constant for the first half of training and decreasing linearly to zero during the second half. The specific architectures for our networks used are detailed in Table \ref{tbl:architecture} in the appendix. We run our experiments for $100n$ epochs, having $n$ domains, as we consider a normal CycleGAN training with two domains to require 200 epochs for adequate results. The fourteen painters dataset, for example, ran 1400 epochs in 220 hours on our nVidia Titan X GPU. Note that pairwise CycleGAN instead would have taken about 2860 hours, or four months. Our code is publicly available at \url{https://github.com/AAnoosheh/ComboGAN}

\subsection{Discussion}
Figure~\ref{fig:alps} shows validation image results for ComboGAN trained on the Alps seasons photos for 400 iterations. ComboGAN did reasonably well converting among the four domains. Looking closely one can notice many examples hide information necessary for the reconstruction process (from training) within them. Many are semantically-meaningful, such as the cloud inversion in the summer images, while some are easy ways to change color back and forth, such as color inversion. Meanwhile in Figure \ref{fig:vanilla_alps} we show results from CycleGAN trained on all six combinations of the four seasons to produce the same images, demonstrating that ComboGAN maintains comparable quality, while only training four networks for 400 epochs instead of CyleGAN's twelve nets for a total of 1200 epochs.

Figure~\ref{fig:paintings} shows randomly-chosen validation images for our fourteen painters dataset. The figure contains translations of a single real image from each artist to every other one. Looking at columns as a whole, one can see common texture behavior and color palettes common to the pieces per artist column. In addition, we have included further real sample artworks from each artist in Figure~\ref{fig:samples} to help give a better impression of what an artist's style is supposed to be. One piece in the translation results which stands out almost immediately is the tenth item under Chagall's column: this image was styled as a completely black-and-white sketch. The datasets gathered did happen to contain a few artworks which were unfinished, preliminary sketches for paintings; this led to the translation model coincidentally choosing to translate Corot's painting to a monochrome pencil/charcoal sketch. Comparison with CycleGAN is not shown as it is computationally infeasible.

\section{Conclusion}
We have shown a novel construction of the CycleGAN model as ComboGAN, which solves the $\theta(n^2)$ scaling issue inherent in current image translation experiments. ComboGAN still maintains the visual richness of CycleGAN without being constrained to two domains. In theory, additional domains can be appended to an existing ComboGAN model by simply creating a new encoder/decoder pair to train alongside a pretrained model.

Though the proposed framework is not restricted to CycleGAN, its formulation can be easily extended to UNIT~\cite{NVIDIA}, for example. The model allows for more modifications, such as encoder-decoder layer sharing, or to add latent-space losses to the representations outputted by the encoders. These were omitted from this work to demonstrate the sole effect of scaling the CycleGAN model and showing it still compares to the original, without introducing scaling-irrelevant adjustments that might improve results on their own.


{\small
\bibliographystyle{ieee}
\bibliography{refs}
}

\newpage
\appendix

\section{Network Architectures}

The network architecture used translation experiments is detailed in Table \ref{tbl:architecture}. We use the following abbreviations for brevity: N=Neurons, K=Kernel size, S=Stride size. The transposed convolutional layer is denoted by DCONV. The residual basic block is denoted as RESBLK.

\begin{table}[hb] \label{tbl:architecture}
\centering
\caption{Layer specifications for Generator (Encoder + Decoder) and Discriminator}
\label{my-label}
\begin{tabular}{cl}
\hline
Layer \#                       & Encoders                                \\ \hline
1                              & CONV-(N64,K7,S1), InstanceNorm, ReLU    \\
2                              & CONV-(N128,K3,S2), InstanceNorm, ReLU   \\
3                              & CONV-(N256,K3,S2), InstanceNorm, ReLU   \\
4                              & RESBLK-(N256,K3,S1), InstanceNorm, ReLU \\
5                              & RESBLK-(N256,K3,S1), InstanceNorm, ReLU \\
6                              & RESBLK-(N256,K3,S1), InstanceNorm, ReLU \\
7                              & RESBLK-(N256,K3,S1), InstanceNorm, ReLU \\ \hline
Layer \#					   & Decoders					             \\ \hline
1                              & RESBLK-(N256,K3,S1), InstanceNorm, ReLU \\
2                              & RESBLK-(N256,K3,S1), InstanceNorm, ReLU \\
3                              & RESBLK-(N256,K3,S1), InstanceNorm, ReLU \\
4                              & RESBLK-(N256,K3,S1), InstanceNorm, ReLU \\
5                              & RESBLK-(N256,K3,S1), InstanceNorm, ReLU \\
6                              & DCONV-(N128,K4,S2), InstanceNorm, ReLU  \\
7                              & DCONV-(N64,K4,S2), InstanceNorm, ReLU   \\
8                              & CONV-(N3,K7,S1), Tanh                   \\ \hline
Layer \#					   & Discriminators				             \\ \hline
1 							   & CONV-(N64,K4,S2), LeakyReLU                \\
2 							   & CONV-(N128,K4,S2), InstanceNorm, LeakyReLU \\
3 							   & CONV-(N256,K4,S2), InstanceNorm, LeakyReLU \\
4 							   & CONV-(N512,K4,S1), InstanceNorm, LeakyReLU \\
5 							   & CONV-(N1,K4,S1)          			     \\ \hline
\end{tabular}
\end{table}

\end{document}